\documentclass{bmvc2k}

\title{How to Train Your Energy-Based Model\\ for Regression}

\addauthor{Fredrik K.~Gustafsson}{fredrik.gustafsson@it.uu.se}{1}
\addauthor{Martin Danelljan}{martin.danelljan@vision.ee.ethz.ch}{2}
\addauthor{Radu Timofte}{radu.timofte@vision.ee.ethz.ch}{2}
\addauthor{Thomas B.~Sch\"on}{thomas.schon@it.uu.se}{1}

\addinstitution{
Department of Information Technology\\
Uppsala University\\
Sweden
}
\addinstitution{
Computer Vision Lab\\
ETH Z\"urich\\
Switzerland
}

\runninghead{F.K. Gustafsson et al.}{How to Train Your EBM for Regression}

\def\eg{\emph{e.g}\bmvaOneDot}

\usepackage{xcolor, soul}
\sethlcolor{pink}

\usepackage{xparse, mathtools, array}
\DeclarePairedDelimiterX{\rvect}[1]{[}{]}{\,\makervect{#1}\,}
\ExplSyntaxOn
\NewDocumentCommand{\makervect}{m}
 {
  \seq_set_split:Nnn \l_tmpa_seq { , } { #1 }
  \begin{matrix}
  \seq_use:Nn \l_tmpa_seq { & }
  \end{matrix}
 }
\ExplSyntaxOff

\usepackage[title]{appendix}

\usepackage{booktabs}
\usepackage{multirow}

\usepackage{algorithm}
\usepackage[noend]{algpseudocode}
\makeatletter
\def\BState{\State\hskip-\ALG@thistlm}
\makeatother
\algnewcommand\Or{\textbf{or}}

\usepackage{amsmath,amsfonts,bm}

\def\eqref#1{equation~\ref{#1}}

\def\1{\bm{1}}

\DeclareMathAlphabet{\mathsfit}{\encodingdefault}{\sfdefault}{m}{sl}
\SetMathAlphabet{\mathsfit}{bold}{\encodingdefault}{\sfdefault}{bx}{n}

\DeclareMathOperator*{\argmax}{arg\,max}

\usepackage{amssymb}

\newcommand{\parsection}[1]{\vspace{2mm}\noindent\textbf{#1}~ }

\usepackage{pgfplots}
    \usepgfplotslibrary{colorbrewer}
    \pgfplotsset{
        cycle list/Dark2,
        cycle multiindex* list={
            mark list*\nextlist
            Dark2\nextlist
        },
    }
\pgfplotsset{compat=1.14}

\begin{document}

\maketitle

\begin{abstract}
Energy-based models (EBMs) have become increasingly popular within computer vision in recent years. While they are commonly employed for generative image modeling, recent work has applied EBMs also for regression tasks, achieving state-of-the-art performance on object detection and visual tracking. Training EBMs is however known to be challenging. While a variety of different techniques have been explored for generative modeling, the application of EBMs to regression is not a well-studied problem. How EBMs should be trained for best possible regression performance is thus currently unclear. We therefore accept the task of providing the first detailed study of this problem. To that end, we propose a simple yet highly effective extension of noise contrastive estimation, and carefully compare its performance to six popular methods from literature on the tasks of 1D regression and object detection. The results of this comparison suggest that our training method should be considered the go-to approach. We also apply our method to the visual tracking task, achieving state-of-the-art performance on five datasets. Notably, our tracker achieves $63.7\%$ AUC on LaSOT and $78.7\%$ Success on TrackingNet. Code is available at \url{https://github.com/fregu856/ebms_regression}. 
\end{abstract}

\section{Introduction}
\label{section:introduction}
Energy-based models (EBMs)~\cite{lecun2006tutorial} have a rich history in machine learning \cite{teh2003energy, bengio2003neural, mnih2005learning, hinton2006unsupervised, osadchy2005synergistic}. An EBM specifies a probability density $p(x; \theta) = e^{f_\theta(x)}/\int e^{f_\theta(x)} dx$ directly via a parameterized scalar function $f_\theta(x)$. By defining $f_\theta(x)$ using a deep neural network (DNN), $p(x; \theta)$ becomes expressive enough to learn practically any density from observed data. EBMs have therefore become increasingly popular within computer vision in recent years, commonly being applied for various generative image modeling tasks \cite{xie2016theory, gao2018learning, nijkamp2019learning, du2019implicit, Grathwohl2020Your, nijkamp2019anatomy, gao2019flow}. 

Recent work \cite{gustafsson2019learning, danelljan2020probabilistic} has also explored conditional EBMs as a general formulation for regression, demonstrating particularly impressive performance on the tasks of object detection \cite{Ren2015FasterRT, law2018cornernet, zhou2019bottom} and visual tracking \cite{SiamRPN++, danelljan2019atom, bhat2019learning}. Regression entails predicting a continuous target~$y$ from an input $x$, given a training set of observed input-target pairs. This was addressed in \cite{gustafsson2019learning, danelljan2020probabilistic} by learning a conditional EBM $p(y | x; \theta)$, capturing the distribution of the target value $y$ given the input $x$. At test time, gradient ascent was then used to maximize $p(y | x; \theta)$ w.r.t.\ $y$, producing highly accurate predictions. Regression is a fundamental problem within computer vision with many additional applications \cite{lathuiliere2019comprehensive, xiao2018simple, yang2019fsa, rothe2016deep, pan2018mean}, which all would benefit from such accurate predictions. In this work, we therefore study the use of EBMs for regression in detail, aiming to further improve its performance and applicability.

While the modeling capacity of EBMs makes them highly attractive for many applications, training EBMs is known to be challenging. This is because the EBM $p(x; \theta) = e^{f_\theta(x)}/\int e^{f_\theta(x)} dx$ involves an intractable integral, complicating the use of standard maximum likelihood (ML) learning. A variety of different techniques have therefore been explored in the generative modeling literature, including alternative estimation methods \cite{gutmann2010noise, gao2019flow, hyvarinen2005estimation, vincent2011connection, song2019generative} and approximations based on Markov chain Monte Carlo (MCMC) \cite{hinton2002training, du2019implicit, nijkamp2019learning, nijkamp2019anatomy}. The application of EBMs for regression is however not a particularly well-studied problem. \cite{gustafsson2019learning, danelljan2020probabilistic} both applied importance sampling to approximate intractable integrals, an approach known to scale poorly with the data dimensionality, and considered no alternative techniques. How EBMs $p(y | x; \theta)$ should be trained for best possible performance on computer vision regression tasks is thus an open question, which we set out to investigate in this work.

\parsection{Contributions}
We propose a simple yet highly effective extension of noise contrastive estimation (NCE) \cite{gutmann2010noise} to train EBMs $p(y | x; \theta)$ for regression tasks. Our proposed method, termed \textit{NCE+}, can be understood as a direct generalization of NCE, accounting for noise in the annotation process. We evaluate NCE+ on illustrative 1D regression problems and on the task of bounding box regression in object detection. We also provide a detailed comparison of NCE+ and \emph{six} popular methods from previous work, the results of which suggest that NCE+ should be considered the go-to training method. Lastly, we apply our proposed NCE+ to the task of visual tracking, achieving state-of-the-art results on \emph{five} common datasets. 

\begin{figure}[t]
    \centering
    \includegraphics[width=0.8925\textwidth]{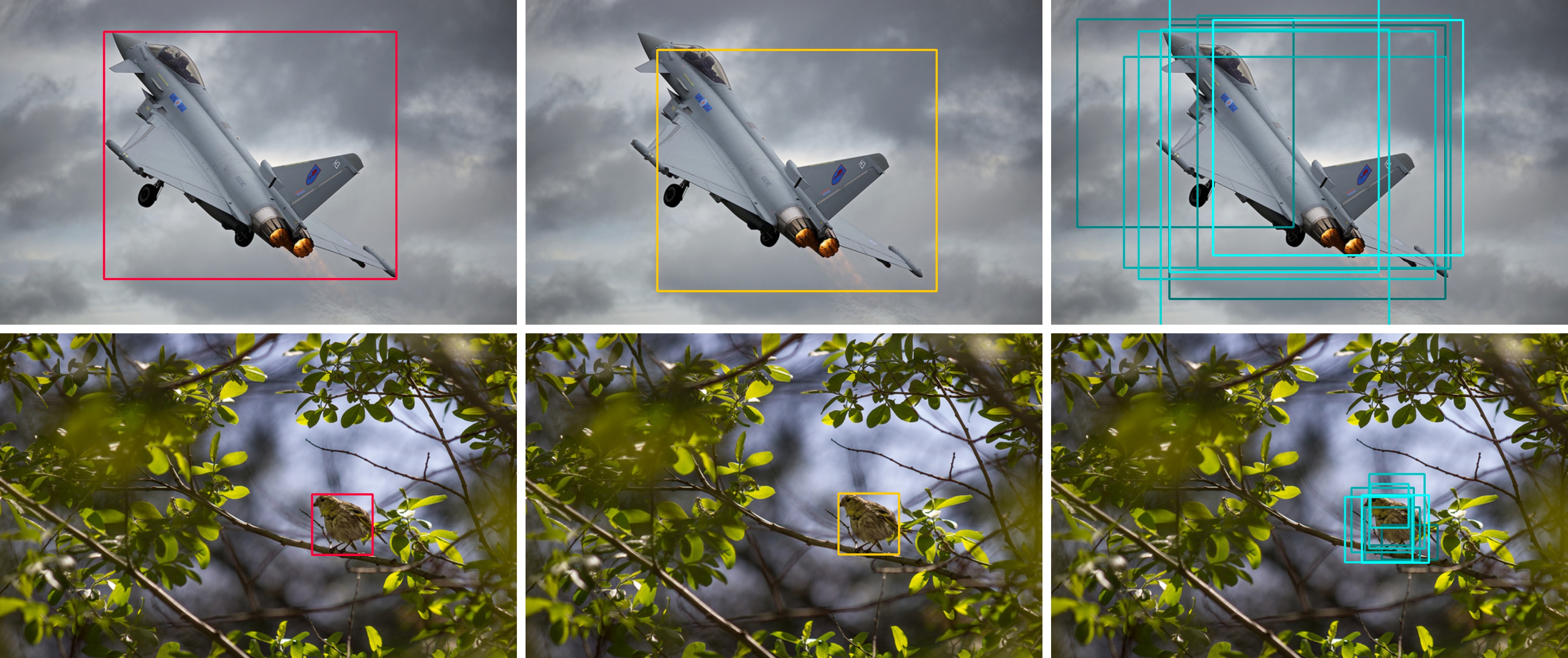}\vspace{-2.0mm}
    \caption{We propose NCE+ to train EBMs $p(y | x; \theta)$ for tasks such as bounding box regression. NCE+ is a highly effective extension of NCE, accounting for noise in the annotation process of real-world datasets. Given a label $y_i$ (red box), the EBM is trained by having to discriminate between $y_i + \nu_i$ (yellow box) and noise samples $\{y^{(i, m)}\}_{m=1}^{M}$ (blue boxes).}\vspace{-3mm}
    \label{fig:overview}
\end{figure}
\section{Energy-Based Models for Regression}
\label{section:ebms}
We study the application of EBMs to important regression tasks in computer vision, using energy-based models of the conditional density $p(y | x)$. Here, we first define the general regression problem and our employed EBM in Section~\ref{section:ebms-problem}. Our prediction strategy based on gradient ascent is then described in Section~\ref{section:ebms-prediction}. Lastly, we discuss the challenges associated with training EBMs, and describe six popular methods from the literature, in Section \ref{section:ebms-training}.

\subsection{Problem \& Model Definition}
\label{section:ebms-problem}
In a supervised regression problem, we are given a training set $\mathcal{D}$ of i.i.d.\ input-target pairs, $\mathcal{D} = \{(x_i, y_i)\}_{i=1}^{N}$, $(x_i, y_i) \sim p(x, y)$. The task is then to learn how to predict a target $y^\star \in \mathcal{Y}$ given a new input $x^\star \in \mathcal{X}$. The target space $\mathcal{Y}$ is continuous, $\mathcal{Y}=\mathbb{R}^K$ for some $K \geq 1$, and the input space $\mathcal{X}$ usually corresponds to the space of images.

As in \cite{gustafsson2019learning, danelljan2020probabilistic}, we address this problem by creating an energy-based model $p(y | x; \theta)$ of the conditional target density $p(y | x)$. To that end, we specify a DNN $f_{\theta}: \mathcal{X} \times \mathcal{Y} \rightarrow \mathbb{R}$ with model parameters $\theta \in \mathbb{R}^P$. This DNN directly maps any input-target pair $(x, y) \in \mathcal{X} \times \mathcal{Y}$ to a scalar $f_{\theta}(x, y) \in \mathbb{R}$. The model $p(y | x; \theta)$ of the conditional target density is then defined as,
\begin{equation}
    p(y | x; \theta) = \frac{e^{f_{\theta}(x, y)}}{Z(x, \theta)}, \qquad Z(x, \theta) = \int e^{f_{\theta}(x, \tilde{y})} d\tilde{y},
\label{eq:ebm_def}
\end{equation}
where the DNN output $f_{\theta}(x, y) \in \mathbb{R}$ is interpreted as the negative energy of the density, and $Z(x, \theta)$ is the input-dependent normalizing partition function. Since $p(y | x; \theta)$ in (\ref{eq:ebm_def}) is directly defined by the DNN $f_{\theta}$, minimal restricting assumptions are put on the true $p(y | x)$. The predictive power of the DNN can thus be fully exploited, enabling learning of, \eg, multi-modal and asymmetric densities directly from data. This expressivity however comes at the cost of $Z(x, \theta)$ being intractable, which complicates evaluating or sampling from $p(y | x; \theta)$.

\subsection{Prediction}
\label{section:ebms-prediction}
At test time, the problem of predicting a target value $y^{\star}$ from an input $x^{\star}$ corresponds to finding a point estimate of the predicted conditional density $p(y | x^\star; \theta)$. The most natural choice is to select the most likely target under the model, $y^\star = \argmax_y p(y | x^\star; \theta) = \argmax_y f_\theta(x^\star, y)$. The prediction $y^\star$ is thus obtained by directly maximizing the DNN scalar output $f_{\theta}(x^\star, y)$ w.r.t.\ $y$, not requiring $Z(x^\star, \theta)$ to be evaluated nor any samples from $p(y | x^\star; \theta)$ to be generated. Following \cite{gustafsson2019learning, danelljan2020probabilistic}, we estimate $y^\star = \argmax_y f_\theta(x^\star, y)$ by performing gradient ascent to refine an initial estimate $\hat{y}$ and find a local maximum of $f_\theta(x^\star, y)$. Starting at $y = \hat{y}$, we thus run $T$ gradient ascent iterations, $y \gets y + \lambda \nabla_{y} f_{\theta}(x^\star, y)$, with step-length $\lambda$. An algorithm for this prediction procedure is found in the supplementary material.

\subsection{Training}
\label{section:ebms-training}
To train the DNN $f_\theta(x, y)$ specifying the EBM (\ref{eq:ebm_def}), different techniques for fitting a density $p(y | x; \theta)$ to observed data $\{(x_i, y_i)\}_{i=1}^{N}$ can be used. In general, the most commonly applied such technique is ML learning, which entails minimizing the negative log-likelihood (NLL), 
\begin{equation}
    -\sum_{i=1}^{N} \log p(y_i | x_i; \theta) = \sum_{i=1}^{N} \log \bigg( \int e^{f_{\theta}(x_i, y)} dy \bigg) - f_{\theta}(x_i, y_i),
\label{eq:ebm_nll}
\end{equation}
w.r.t.\ the parameters $\theta$. The integral in (\ref{eq:ebm_nll}) is however intractable, and exact evaluation of the NLL is thus not possible. \cite{gustafsson2019learning,danelljan2020probabilistic} employed importance sampling to approximate such intractable integrals, obtaining state-of-the-art performance on object detection and visual tracking. Recent work \cite{gao2019flow, saremi2018deep, li2019annealed, du2019implicit, nijkamp2019anatomy, Grathwohl2020Your} on generative image modeling has however applied a variety of different training methods not considered in \cite{gustafsson2019learning,danelljan2020probabilistic}, including the ML learning alternatives NCE~\cite{gutmann2010noise} and score matching~\cite{hyvarinen2005estimation}. How we should train the DNN $f_\theta$ to obtain best possible regression performance is thus unclear. In this work, we therefore carefully compare our proposed method to six popular training methods from the literature.

\parsection{ML with Importance Sampling (ML-IS)}
A straightforward training method is proposed in \cite{gustafsson2019learning}, which we term \textit{ML with Importance Sampling (ML-IS)}. Using ML-IS, \cite{gustafsson2019learning} successfully applied the EBM (\ref{eq:ebm_def}) to the regression tasks of object detection, visual tracking, age estimation and head-pose estimation. In ML-IS, the DNN $f_\theta$ is trained by directly minimizing the NLL (\ref{eq:ebm_nll}) w.r.t.\ $\theta$, using importance sampling to approximate the intractable integral,
\begin{equation}
    -\log p(y_i| x_i;\theta) \approx \log \bigg(\frac{1}{M} \sum_{m=1}^{M} \frac{e^{f_{\theta}(x_i, y^{(i, m)})}}{q(y^{(i, m)}|y_i)} \bigg) - f_{\theta}(x_i, y_i).
\label{eq:mle-is_loss_term}
\end{equation}
Here, $\{y^{(i, m)}\}_{m=1}^{M}$ are $M$ samples drawn from a proposal distribution $q(y|y_i)$ that depends on the ground truth target $y_i$. In \cite{gustafsson2019learning}, $q(y|y_i)$ is set to a mixture of $K$ Gaussians centered at $y_i$,
\begin{equation}
    q(y|y_i) = \frac{1}{K} \sum_{k=1}^{K} \mathcal{N}(y; y_i, \sigma_{k}^{2}I).
\label{eq:mle-is_proposal}
\end{equation}
The loss $J(\theta)$ is obtained by averaging over all pairs $\{(x_i,y_i)\}_{i=1}^{n}$ in the current mini-batch,
\begin{equation}
    J(\theta) = \frac{1}{n} \sum_{i = 1}^{n} \log \bigg(\frac{1}{M} \sum_{m=1}^{M} \frac{e^{f_{\theta}(x_i, y^{(i, m)})}}{q(y^{(i, m)}|y_i)} \bigg) - f_{\theta}(x_i, y_i).
\label{eq:mle-is_loss}
\end{equation}

\parsection{KL Divergence with Importance Sampling (KLD-IS)}
Instead of minimizing the NLL~(\ref{eq:ebm_nll}), \cite{danelljan2020probabilistic} considers the Kullback-Leibler (KL) divergence $D_\mathrm{KL}(p(y | y_i) \parallel p(y | x_i; \theta))$ between the EBM $p(y | x_i; \theta)$ and an assumed density $p(y | y_i)$ of the true target $y$ given the label $y_i$. The density $p(y | y_i)$ models noise in the annotation process of our given training set $\mathcal{D} = \{(x_i, y_i)\}_{i=1}^{N}$. In \cite{danelljan2020probabilistic}, $p(y | y_i) = \mathcal{N}(y; y_i, \sigma^2 I)$, where $\sigma$ is a hyperparameter. As shown in \cite{danelljan2020probabilistic},
\begin{equation}
    D_\mathrm{KL}(p(y | y_i) \parallel p(y | x_i; \theta)) = \log \bigg( \int e^{f_{\theta}(x_i, y)} dy \bigg) - \int f_{\theta}(x_i, y) p(y | y_i) dy + C,
\label{eq:kld-is_kldiv}
\end{equation}
where $C$ is a constant that does not depend on $\theta$. \cite{danelljan2020probabilistic} approximates the integrals in (\ref{eq:kld-is_kldiv}) using importance sampling, employing the ML-IS proposal $q(y|y_i)$ in (\ref{eq:mle-is_proposal}). By then averaging over all pairs $\{(x_i,y_i)\}_{i=1}^{n}$ in the current mini-batch, the loss $J(\theta)$ used to train $f_{\theta}$ is obtained as,
\begin{equation}
    J(\theta) = \frac{1}{n} \sum_{i = 1}^{n} \log \bigg(\frac{1}{M} \sum_{m=1}^{M} \frac{e^{f_{\theta}(x_i, y^{(i, m)})}}{q(y^{(i, m)}|y_i)} \bigg) - \frac{1}{M} \sum_{m=1}^{M} f_{\theta}(x_i, y^{(i, m)}) \frac{p(y^{(i, m)}|y_i)}{q(y^{(i, m)}|y_i)},
\label{eq:kld-is_loss}
\end{equation}
where $\{y^{(i, m)}\}_{m=1}^{M}$ are $M$ samples drawn from the proposal $q(y|y_i)$. We term this training method \textit{KL Divergence with Importance Sampling (KLD-IS)}. When applied to visual tracking in \cite{danelljan2020probabilistic}, KLD-IS outperformed ML-IS and set a new state-of-the-art.

\parsection{ML with MCMC (ML-MCMC)}
To minimize the NLL (\ref{eq:ebm_nll}) w.r.t.\ the parameters $\theta$, the following identity for the expression of the gradient $\nabla_{\theta} -\log p(y_i | x_i; \theta)$ can be utilized \cite{lecun2006tutorial},
\begin{equation}
    \nabla_{\theta} -\log p(y_i | x_i; \theta) = \mathbb{E}_{p(y | x_i; \theta)}\bigg[ \nabla_{\theta} f_{\theta}(x_i, y) \bigg] - \nabla_{\theta} f_{\theta}(x_i, y_i).    
\label{eq:mle-mcmc_gradient} 
\end{equation} 
The expectation in (\ref{eq:mle-mcmc_gradient}) is then approximated using samples $\{y^{(i, m)}\}_{m=1}^{M}$ drawn from $p(y | x_i; \theta)$, \textit{i.e.} from the EBM itself. To obtain each sample $y^{(i, m)} \sim p(y | x_i; \theta)$, MCMC is used. Specifically, we follow recent work \cite{xie2016theory, gao2018learning, du2019implicit, nijkamp2019learning, nijkamp2019anatomy, Grathwohl2020Your} on generative image modeling and run $L \geq 1$ steps of Langevin dynamics~\cite{welling2011bayesian}. Starting at $y_{(0)}$, we thus update $y_{(l)}$ according to,
\begin{equation}
    y_{(l+1)} = y_{(l)} + \frac{\alpha^2}{2} \nabla_{y} f_{\theta}(x_i, y_{(l)}) + \alpha \epsilon_{l}, \quad \epsilon_{l} \sim \mathcal{N}(0, I),
\label{eq:mle-mcmc_langevin_dynamics} 
\end{equation}
and set $y^{(i, m)} = y_{(L)}$. Here, $\alpha > 0$ is a small constant step-length. Following the principle of contrastive divergence~\cite{lecun2006tutorial, hinton2002training, teh2003energy}, we start the Markov chain~(\ref{eq:mle-mcmc_langevin_dynamics}) at the ground truth target, $y_{(0)} = y_i$. By approximating (\ref{eq:mle-mcmc_gradient}) with the samples $\{y^{(i, m)}\}_{m=1}^{M}$, and by averaging over all pairs $\{(x_i,y_i)\}_{i=1}^{n}$ in the current mini-batch, the loss $J(\theta)$ used to train the DNN $f_{\theta}$ is obtained as,
\begin{equation}
    J(\theta) = \frac{1}{n} \sum_{i = 1}^{n} \bigg( \frac{1}{M} \sum_{m=1}^{M} f_{\theta}(x_i, y^{(i, m)}) \bigg) - f_{\theta}(x_i, y_i).   
\label{eq:mle-mcmc_loss} 
\end{equation}
We term this specific training method \textit{ML with MCMC (ML-MCMC)}.

\parsection{Noise Contrastive Estimation (NCE)}
As an alternative to ML learning, Gutmann and Hyv{\"a}rinen proposed NCE~\cite{gutmann2010noise} for estimating unnormalized parametric models. NCE entails generating samples from some noise distribution $p_{N}$, and learning to discriminate between these noise samples and observed data examples. It has recently been applied to generative image modeling with EBMs \cite{gao2019flow}, and the NCE loss is also utilized in various frameworks for self-supervised learning \cite{hjelm2018learning, bachman2019learning, chen2020simple}. Moreover, NCE has been applied to train EBMs for supervised \emph{classification} tasks within language modeling \cite{mnih2012fast, mikolov2013distributed, jozefowicz2016exploring, ma2018noise}, where the target space $\mathcal{Y}$ is a large but finite set of possible labels. We adopt NCE for regression by using a noise distribution $p_{N}(y|y_i)$ of the same form as the ML-IS proposal in~(\ref{eq:mle-is_proposal}),
\begin{equation}
    p_{N}(y|y_i) = \frac{1}{K} \sum_{k=1}^{K} \mathcal{N}(y; y_i, \sigma_{k}^{2}I),
\label{eq:nce_noise}
\end{equation}
and by employing the ranking NCE objective~\cite{jozefowicz2016exploring}, as described in \cite{ma2018noise}. We choose ranking NCE over the binary objective since it is consistent under a weaker assumption \cite{ma2018noise}. We thus define $y^{(i, 0)} \triangleq y_i$, and train the DNN $f_\theta$ by minimizing the following loss,
\begin{equation}
    J(\theta) = - \frac{1}{n} \sum_{i = 1}^{n} \log \frac{\exp \big\{f_{\theta}(x_i, y^{(i, 0)}) - \log p_{N}(y^{(i, 0)} | y_i)\big\}}{\sum_{m=0}^{M} \exp \big\{f_{\theta}(x_i, y^{(i, m)}) - \log p_{N}(y^{(i, m)} | y_i)\big\}},   
\label{eq:nce_loss} 
\end{equation}
where $\{y^{(i, m)}\}_{m=1}^{M}$ are $M$ noise samples drawn from $p_{N}(y|y_i)$ in (\ref{eq:nce_noise}).

\parsection{Score Matching (SM)}
Another alternative estimation method is score matching (SM), as proposed by Hyv{\"a}rinen~\cite{hyvarinen2005estimation} and further studied for supervised problems in \cite{sasaki2018neural}. The method focuses on the \emph{score} of $p(y | x; \theta)$, defined as $\nabla_{y} \log p(y | x; \theta) = \nabla_{y} f_{\theta}(x, y)$, aiming for it to approximate the score of the true target density $p(y | x)$. Note that the EBM score $\nabla_{y} f_{\theta}(x, y)$ does not depend on the intractable $Z(x, \theta)$. SM was applied to simple conditional density estimation problems in \cite{sasaki2018neural}, using a combination of feed-forward networks and reproducing kernels to specify the EBM. Following \cite{sasaki2018neural}, we train the DNN $f_\theta$ by minimizing the loss, 
\begin{equation}
    J(\theta) = \frac{1}{n} \sum_{i = 1}^{n} \mathrm{tr} \big( \nabla^{2}_{y} f_{\theta}(x_i, y_i) \big) + \frac{1}{2} \big\| \nabla_{y} f_{\theta}(x_i, y_i) \big\|^2_2,   
\label{eq:sm_loss} 
\end{equation}
where only the diagonal of $\nabla^{2}_{y} f_{\theta}(x_i, y_i)$ actually is needed to compute the first term.

\parsection{Denoising Score Matching (DSM)}
By modifying the SM objective, denoising score matching (DSM) was proposed by Vincent~\cite{vincent2011connection}. DSM does not require computation of any second derivatives, improving its scalability to high-dimensional data. The method entails employing SM on noise-corrupted data points. Recently, DSM has been successfully applied to generative image modeling \cite{saremi2018deep, song2019generative, li2019annealed}. DSM was also extended to train EBMs of conditional densities in \cite{khemakhem2020ice}, where it was applied to a transfer learning problem. Following \cite{khemakhem2020ice}, we use a Gaussian noise distribution and train the DNN $f_\theta$ by minimizing the loss,
\begin{equation}
    J(\theta) = \frac{1}{n} \sum_{i = 1}^{n} \frac{1}{M} \sum_{m = 1}^{M} \bigg\| \nabla_{y} f_{\theta}(x_i, \tilde{y}^{(i, m)}) + \frac{\tilde{y}^{(i, m)} - y_i}{\sigma^{2}} \bigg\|^2_2,
\label{eq:dsm_loss} 
\end{equation}
where $\{\tilde{y}^{(i, m)}\}_{m=1}^{M}$ are $M$ samples drawn from the noise distribution $p_{\sigma}(\tilde{y}|y_i) = \mathcal{N}(\tilde{y}; y_i, \sigma^{2}I)$.
\section{Proposed Training Method}
\label{section:method}
To train the DNN $f_\theta$ specifying our EBM $p(y | x; \theta)$ in (\ref{eq:ebm_def}), we propose a \emph{simple yet highly effective} extension of NCE~\cite{gutmann2010noise}. Motivated by the improved performance of KLD-IS compared to ML-IS on visual tracking \cite{danelljan2020probabilistic}, we extend NCE with the capability to model annotation noise. To that end, we adopt the standard NCE noise distribution $p_{N}$ (\ref{eq:nce_noise}) and loss (\ref{eq:nce_loss}), but instead of defining $y^{(i, 0)} \triangleq y_i$, we sample $\nu_i \sim p_{\beta}(y)$ and define $y^{(i, 0)} \triangleq y_i + \nu_i$. The distribution $p_{\beta}$ is a zero-centered version of $p_{N}$ in which $\{\sigma_{k}\}_{k=1}^{K}$ are scaled with $\beta > 0$,
\begin{equation}
    p_{N}(y|y_i) = \frac{1}{K} \sum_{k=1}^{K} \mathcal{N}(y; y_i, \sigma_{k}^{2}I), \qquad p_{\beta}(y) = \frac{1}{K} \sum_{k=1}^{K} \mathcal{N}(y; 0, \beta \sigma_{k}^{2}I).
\label{eq:nce_noise_}
\end{equation}
Instead of training the DNN $f_\theta$ by learning to discriminate between noise samples $\{y^{(i, m)}\}_{m=1}^{M}$ and the label $y_i$, it thus has to discriminate between the samples $\{y^{(i, m)}\}_{m=1}^{M}$ and $y_i + \nu_i$. Examples of $y_i + \nu_i$ and $\{y^{(i, m)}\}_{m=1}^{M}$ in the task of bounding box regression are visualized in Figure~\ref{fig:overview}. Similar to KLD-IS, in which an assumed density of the true target value $y$ given $y_i$ is employed, our approach thus accounts for possible noise and inaccuracies in the provided label $y_i$. Specifically, our proposed training method entails sampling $\{y^{(i, m)}\}_{m=1}^{M} \sim p_{N}(y|y_i)$ and $\nu_i \sim p_{\beta}(y)$, setting $y^{(i, 0)} \triangleq y_i + \nu_i$, and minimizing the following loss,
\begin{equation}
    J(\theta) = - \frac{1}{n} \sum_{i = 1}^{n} \log \frac{\exp \big\{f_{\theta}(x_i, y^{(i, 0)}) - \log p_{N}(y^{(i, 0)} | y_i)\big\}}{\sum_{m=0}^{M} \exp \big\{f_{\theta}(x_i, y^{(i, m)}) - \log p_{N}(y^{(i, m)} | y_i)\big\}}.   
\label{eq:nce_loss_} 
\end{equation}
As $\beta \rightarrow 0$, samples $\nu_i \sim p_{\beta}(y)$ will concentrate increasingly close to zero, and the standard NCE method is in practice recovered. Our proposed training method can thus be understood as a direct generalization of NCE. Compared to NCE, our method adds no significant training cost and requires tuning of a single additional hyperparameter $\beta$. A value for $\beta$ is selected in a simple two-step procedure. First, we fix $y^{(i, 0)} = y_i$ and select the standard deviations $\{\sigma_{k}\}_{k=1}^{K}$ based on validation set performance, just as in NCE. We then fix $\{\sigma_{k}\}_{k=1}^{K}$ and vary $\beta$ to find the value corresponding to maximum validation performance. Typically, we start this ablation with $\beta = 0.1$. We term our proposed training method \textit{NCE+}.
\section{Comparison of Training Methods}
\label{section:comparison}
We provide a detailed comparison of the six training methods from Section~\ref{section:ebms-training} and our proposed NCE+. To that end, we perform extensive experiments on 1D regression (Section~\ref{section:comparison-1DRegression}) and object detection (Section~\ref{section:comparison-ObjectDetection}). Our findings are summarized in Section~\ref{section:comparison-Observations}. All experiments are implemented in PyTorch~\cite{paszke2019pytorch} and the code is publically available. For both tasks, further details and results are also provided in the supplementary material.

\subsection{1D Regression Experiments}
\label{section:comparison-1DRegression}
We first perform experiments on illustrative 1D regression problems. The DNN $f_\theta(x, y)$ is here a simple feed-forward network, taking $x \in \mathbb{R}$ and $y \in \mathbb{R}$ as inputs. We employ two synthetic datasets, and evaluate the training methods by how well the learned model $p(y | x; \theta)$ (\ref{eq:ebm_def}) approximates the known ground truth $p(y | x)$, as measured by the KL divergence $D_\mathrm{KL}$.

\parsection{Results}
A comparison of all seven training methods in terms of $D_\mathrm{KL}$ and training cost (seconds per epoch) is found in Table~\ref{tab:comparison_1dregression}. For ML-MCMC, we include results for $L \in \{1,16,256\}$ Langevin steps (\ref{eq:mle-mcmc_langevin_dynamics}). We observe that ML-IS, KLD-IS, NCE and NCE+ clearly have the best performance. While ML-MCMC is relatively close in terms of $D_\mathrm{KL}$ for $L=256$, this comes at the expense of a massive increase in training cost. DSM outperforms SM in terms of both metrics, but is not close to the top-performing methods. The four best methods are further compared in Figure~\ref{fig:comparison_1dregression_performance_vs_M}, showing $D_\mathrm{KL}$ as a function of $M$. Here, we observe that NCE and NCE+ significantly outperform ML-IS and KLD-IS for small number of samples $M$.

\begin{table}[t]
\centering
	\resizebox{1.0\textwidth}{!}{%

\begin{tabular}{l@{\hspace{0.35cm}}ccccccccc}
\toprule
 &ML-IS &ML-MCMC-1 &ML-MCMC-16 &ML-MCMC-256 &KLD-IS &NCE &SM &DSM &$\textbf{NCE+}$\\
\midrule
$D_\mathrm{KL}$ $\downarrow$         &\textbf{0.062} &0.865 &0.449 &0.106 &0.088 &0.068 &0.781 &0.395 &0.066\\
Training Cost $\downarrow$  &\textbf{0.44} &0.54 &2.41 &30.8 &\textbf{0.44} &0.45 &0.60 &0.47 &0.46\\
\bottomrule
\end{tabular}
	}\vspace{-3.5mm}
	\caption{Comparison of training methods for the illustrative 1D regression experiments.}\vspace{-1.5mm}
	\label{tab:comparison_1dregression}
\end{table} 

\begin{figure}%
    \begin{minipage}{0.475\textwidth}%
        \centering
        \begin{tikzpicture}[scale=0.655, baseline]
            \begin{axis}[
                xmode=log,
                log ticks with fixed point,
                xlabel={Number of samples $M$},
                ylabel={$D_\mathrm{KL}$},
                xtick={1, 4, 16, 64, 256, 1024},
                xticklabels={1, 4, 16, 64, 256, 1024},
                legend pos=north east,
                ymajorgrids=true,
                grid style=dashed,
                every axis plot/.append style={thick},
            ]
            \addplot
             plot [error bars/.cd, y dir = both, y explicit]
             table[row sep=crcr, x index=0, y index=1]{
            1 3.1431\\
            4 1.449767\\
            16 0.13682405\\
            64 0.07923745\\ 
            256 0.06401\\
            1024 0.062326016666666664\\
            };
            \addlegendentry{ML-IS}
            
            \addplot
             plot [error bars/.cd, y dir = both, y explicit]
             table[row sep=crcr, x index=0, y index=1]{
            1 2.512175\\
            4 0.9612085\\
            16 0.4151015\\ 
            64 0.21802749999999999\\
            256 0.12049625\\ 
            1024 0.08826348333333334\\
            };
            \addlegendentry{KLD-IS}
            
            \addplot
             plot [error bars/.cd, y dir = both, y explicit]
             table[row sep=crcr, x index=0, y index=1]{
            1 0.48795700000000003\\
            4 0.222217\\
            16 0.12022195\\
            64 0.07445745\\
            256 0.0630213\\ 
            1024 0.06818718333333332\\
            };
            \addlegendentry{NCE}
            
            \addplot
             plot [error bars/.cd, y dir = both, y explicit]
             table[row sep=crcr, x index=0, y index=1]{
            1 0.41710800000000003\\
            4 0.1957025\\ 
            16 0.10278555\\
            64 0.0797574\\ 
            256 0.0631705\\ 
            1024 0.06565895\\
            };
            \addlegendentry{\textbf{NCE+}}
            
            \end{axis}
        \end{tikzpicture}
        \caption{Detailed comparison of the top-performing methods for the illustrative 1D regression experiments. NCE and NCE+ here demonstrate clear superior performance for small number of samples $M$.}\vspace{-3mm}
        \label{fig:comparison_1dregression_performance_vs_M}%
    \end{minipage}
    \quad
    \begin{minipage}{0.475\textwidth}%
            \begin{tikzpicture}[scale=0.655, baseline]
                \begin{axis}[
                    xmode=log,
                    log ticks with fixed point,
                    xlabel={Number of samples $M$},
                    ylabel={AP (\%)},
                    ymin=32.5, ymax=39.73,
                    xtick={1, 2, 4, 8, 16, 32, 64, 128},
                    ytick={33, 34, 35, 36, 37, 38, 39},
                    legend pos=south east,
                    ymajorgrids=true,
                    grid style=dashed,
                    every axis plot/.append style={thick},
                ]
                \addplot
                 plot [error bars/.cd, y dir = both, y explicit]
                 table[row sep=crcr, x index=0, y index=1]{
                8 33.06\\
                16 36.33\\
                32 37.68\\
                64 38.96\\
                128 39.11\\
                };
                \addlegendentry{ML-IS}
                
                \addplot
                 plot [error bars/.cd, y dir = both, y explicit]
                 table[row sep=crcr, x index=0, y index=1]{
                4 32.77\\
                8 38.32\\
                16 38.81\\
                32 39.10\\
                64 39.21\\
                128 39.37\\
                };
                \addlegendentry{KLD-IS}
                
                \addplot
                 plot [error bars/.cd, y dir = both, y explicit]
                 table[row sep=crcr, x index=0, y index=1]{
                1 37.82\\
                2 38.16\\
                4 38.46\\
                8 38.75\\
                16 38.81\\
                32 38.91\\
                64 39.07\\
                128 39.17\\
                };
                \addlegendentry{NCE}
                
                \addplot
                 plot [error bars/.cd, y dir = both, y explicit]
                 table[row sep=crcr, x index=0, y index=1]{
                1 37.93\\
                2 38.27\\
                4 38.53\\
                8 38.82\\
                16 39.01\\
                32 39.18\\
                64 39.29\\
                128 39.36\\
                };
                \addlegendentry{\textbf{NCE+}}
                
                \end{axis}
            \end{tikzpicture}
        \caption{Detailed comparison of the top-performing methods for object detection, on the \textit{2017 val} split of COCO~\cite{lin2014microsoft}. Missing values for ML-IS and KLD-IS correspond to failed training due to numerical issues.}\vspace{-3mm}
        \label{fig:comparison_detection_AP_vs_M}%
    \end{minipage}%
\end{figure}%

\subsection{Object Detection Experiments}
\label{section:comparison-ObjectDetection}
Next, we evaluate the methods on the task of bounding box regression in object detection. We employ an identical network architecture for $f_{\theta}(x, y)$ as in \cite{gustafsson2019learning}. An extra network branch, consisting of three fully-connected layers with parameters $\theta$, is thus added onto a pre-trained and fixed FPN Faster-RCNN detector~\cite{lin2017feature}. Given an image $x$ and bounding box $y \in \mathbb{R}^{4}$, the image is first processed by the detector backbone network (ResNet50-FPN), outputting image features $h_{1}(x)$. Using a differentiable PrRoiPool~\cite{jiang2018acquisition} layer, $h_{1}(x)$ is then pooled to extract features $h_{2}(x, y)$. Finally, $h_{2}(x, y)$ is processed by the added network branch, outputting $f_{\theta}(x, y) \in \mathbb{R}$. As in \cite{gustafsson2019learning}, predictions $y^{\star}$ are produced by performing guided NMS~\cite{jiang2018acquisition} followed by gradient-based refinement (Section~\ref{section:ebms-prediction}), taking the Faster-RCNN detections as initial estimates $\hat{y}$. Experiments are performed on the large-scale COCO dataset~\cite{lin2014microsoft}. We use the \textit{2017 train} split ($\approx$ 118\thinspace000 images) for training, the \textit{2017 val} split ($\approx$ 5\thinspace000 images) for setting hyperparameters, and report results on the \emph{2017 test-dev} split ($\approx$~20\thinspace000 images). The standard COCO metrics AP, AP$_\text{50}$ and AP$_\text{75}$ are used, where AP is the primary metric.


\parsection{Results}
A comparison of the training methods in terms of the COCO metrics and training cost (seconds per iteration) is found in Table~\ref{tab:comparison_object_detection}. Since DSM clearly outperformed SM in the 1D regression experiments, we here only include DSM. For ML-MCMC, results for $L \in \{1,4,8\}$ are included. We observe that ML-IS, KLD-IS, NCE and NCE+ clearly have the best performance. In terms of the COCO metrics, NCE+ outperforms NCE and all other methods. ML-IS is also outperformed by KLD-IS. The four top-performing methods are further compared in Figure~\ref{fig:comparison_detection_AP_vs_M}, in terms of AP as a function of the number of samples $M$. NCE and NCE+ here demonstrate clear superior performance for small values of $M$, and do not experience numerical issues even for $M\!=\!1$. KLD-IS improves this robustness compared ML-IS, but is not close to matching NCE or NCE+. In terms of training cost, the four top-performing methods are virtually identical. For ML-IS, \eg, we observe in Figure~\ref{fig:training_cost_vs_M} that setting $M\!=\!1$ decreases the training cost with $23\%$ compared to the standard case of $M\!=\!128$.

\parsection{Analysis of NCE+ Hyperparameters}
How the value of $\beta > 0$ in $p_{\beta}$ (\ref{eq:nce_noise_}) affects validation performance is studied in Figure~\ref{fig:effect_of_beta}. Here, we observe that quite a large range of values improve the performance compared to the NCE baseline ($\beta \rightarrow 0$), before it eventually degrades for $\beta \gtrsim 0.3$. We also observe that the performance is optimized for $\beta = 0.1$. In Figure~\ref{fig:effect_of_beta}, the standard deviations $\{\sigma_{k}\}_{k=1}^{K}$ in $p_{N}$, $p_{\beta}$ (\ref{eq:nce_noise_}) are set to $\{0.075, 0.15, 0.3\}$. These values are selected in an initial step based on an ablation study for NCE, which is found in Table~\ref{tab:detection_ablation_nce}.

\begin{table}[t]
\centering
\resizebox{1.0\textwidth}{!}{%

\begin{tabular}{l@{\hspace{0.5cm}}cccccccc}
\toprule
 &ML-IS &ML-MCMC-1 &ML-MCMC-4 &ML-MCMC-8 &KLD-IS &NCE& DSM &\textbf{NCE+}\\
\midrule
AP (\%) $\uparrow$             &39.4 &36.4 &36.4 &36.4 &39.6 &39.5 &36.3 &\textbf{39.7}\\
AP$_\text{50} (\%)$ $\uparrow$ &58.6 &57.9 &57.9 &58.0 &58.6 &58.6 &57.9 &\textbf{58.7}\\
AP$_\text{75} (\%)$ $\uparrow$ &42.1 &38.8 &39.0 &39.0 &42.6 &42.4 &38.9 &\textbf{42.7}\\
Training Cost $\downarrow$     &1.03 &2.47 &7.05 &13.3 &\textbf{1.02} &1.04 &3.84 &1.09\\
\bottomrule
\end{tabular}
	}\vspace{-3.5mm}
\caption{Comparison of training methods for the object detection experiments, on the \textit{2017 test-dev} split of COCO~\cite{lin2014microsoft}. Our proposed NCE+ achieves the best performance.}\vspace{-3.0mm}
	\label{tab:comparison_object_detection}
\end{table}




\subsection{Discussion}%
\label{section:comparison-Observations}
The results on both set of experiments are highly consistent. First of all, ML-IS, KLD-IS, NCE and NCE+ are by far the top-performing training methods. ML-MCMC, the method commonly employed for generative image modeling in recent years, does not come close to matching these top-performing methods, especially not given similar computational budgets. When studying the performance as a function of the number of samples $M$, NCE and NCE+ are the superior methods by a significant margin. In particular, this study demonstrates that the NCE and NCE+ losses are  numerically more stable than those of ML-IS and KLD-IS. In the 1D regression problems, which employ synthetic datasets without any annotation noise, NCE and NCE+ have virtually identical performance. In the object detection experiments however, where we employ real-world datasets, NCE+ consistently improves the NCE performance. On object detection, NCE+ also improves or matches the performance of KLD-IS, which explicitly models annotation noise and outperforms ML-IS. Overall, the results of the comparison suggest that our proposed NCE+ should be considered the go-to training method.


\begin{figure}%
    \begin{minipage}{0.475\textwidth}%
        \begin{tikzpicture}[scale=0.655, baseline]
            \begin{axis}[
                xmode=log,
                log ticks with fixed point,
                xlabel={Number of samples $M$},
                ylabel={Training Cost},
                xtick={1, 2, 4, 8, 16, 32, 64, 128},
                legend pos=south east,
                ymajorgrids=true,
                grid style=dashed,
                every axis plot/.append style={thick},
            ]
            \addplot
             plot [error bars/.cd, y dir = both, y explicit]
             table[row sep=crcr, x index=0, y index=1]{
            1 0.793625\\
            2 0.7975\\
            4 0.79825\\
            8 0.8035\\
            16 0.8285\\
            32 0.848875\\
            64 0.90075\\
            128 1.03336\\
            };
            
            \end{axis}
        \end{tikzpicture}
        \caption{Effect of the number of samples $M$ on training cost (seconds per iteration), for ML-IS on object detection.}\vspace{-1.5mm}
        \label{fig:training_cost_vs_M}%
    \end{minipage}%
    \quad
    \begin{minipage}{0.475\textwidth}%
        \centering
        \begin{tikzpicture}[scale=0.655, baseline]
            \begin{axis}[
                xlabel={$\beta$},
                ylabel={AP (\%)},
                xtick={0, 0.1, 0.2, 0.3, 0.4, 0.5, 0.6, 0.7, 0.8},
                legend pos=south east,
                ymajorgrids=true,
                grid style=dashed,
                every axis plot/.append style={thick},
            ]
            \addplot
             plot [error bars/.cd, y dir = both, y explicit]
             table[row sep=crcr, x index=0, y index=1]{
            0 39.17\\
            0.025 39.20\\ 
            0.05 39.27\\
            0.1 39.36\\
            0.15 39.32\\
            0.2 39.22\\
            0.4 39.07\\
            0.8 38.34\\
            };
            
            \end{axis}
        \end{tikzpicture}
        \caption{Effect of the NCE+ hyperparameter $\beta$ on object detection performance ($\uparrow$), on the \textit{2017 val} split of COCO~\cite{lin2014microsoft}.}\vspace{-1.5mm}
        \label{fig:effect_of_beta}%
    \end{minipage}
\end{figure}%

\begin{table}
\centering
\resizebox{1.0\textwidth}{!}{%
		\begin{tabular}{l@{\hspace{0.5cm}}ccccc}
\toprule
$\{\sigma_{k}\}_{k=1}^{3}$                       &\{0.0125, 0.025, 0.05\} &\{0.025, 0.05, 0.1\} &\{0.05, 0.1, 0.2\} &\{0.075, 0.15, 0.3\} &\{0.1, 0.2, 0.4\}\\
\midrule
AP (\%) $\uparrow$            &38.58 &38.95 &39.12 &\textbf{39.17} &39.05\\
\bottomrule
\end{tabular}
	}\vspace{-3.5mm}
\caption{Ablation study for NCE, on the \textit{2017 val} split of COCO~\cite{lin2014microsoft}.}\vspace{-3.0mm}
	\label{tab:detection_ablation_nce}
\end{table}

\section{Visual Tracking Experiments}
\label{section: experiments}
Lastly, we apply our proposed NCE+ to the task of visual tracking. Specifically, we consider \emph{generic visual object tracking}, which entails estimating the bounding box $y \in \mathbb{R}^{4}$ of a target object in every frame of a video. The target object does not belong to any pre-specified class, but is instead defined by a given bounding box in the initial video frame. We compare the performance both to NCE and KLD-IS, and to state-of-the-art trackers. Code and trained models are available at \cite{pytracking}. Further details are also found in the supplementary material.

\parsection{Tracking Approach}
We base our tracker on the recent DiMP~\cite{bhat2019learning} and PrDiMP~\cite{danelljan2020probabilistic}. The target object is thus first coarsely localized in the current video frame via 2D image-coordinate regression of its center point, emphasizing robustness over accuracy. Then, the full bounding box $y \in \mathbb{R}^{4}$ of the target is accurately regressed by gradient-based refinement (Section~\ref{section:ebms-prediction}). The two stages employ separate network branches which are trained jointly end-to-end. As a strong baseline, we combine the DiMP method for center point regression with the PrDiMP bounding box regression approach. We term this resulting tracker \textit{DiMP-KLD-IS}. By also modifying common training parameters (batch size, data augmentation etc.), DiMP-KLD-IS significantly outperforms both DiMP and PrDiMP. Our proposed tracker, termed \textit{DiMP-NCE+}, is then obtained simply by using NCE+ instead of KLD-IS to train the bounding box regression branch. In both cases, the number of samples $M=128$. As in \cite{bhat2019learning, danelljan2020probabilistic}, the training splits of TrackingNet~\cite{muller2018trackingnet}, LaSOT~\cite{fan2019lasot}, GOT-10k~\cite{huang2019got} and COCO~\cite{lin2014microsoft} are used for training. Similar to PrDiMP, our DiMP-NCE+ tracker runs at about 30 FPS on a single GPU.

\parsection{Results}
We evaluate DiMP-NCE+ on five commonly used tracking datasets. Tracking-Net~\cite{muller2018trackingnet} is a large-scale dataset containing videos sampled from YouTube. Results are reported on its test set of $511$ videos. We also evaluate on the LaSOT~\cite{fan2019lasot} test set, containing $280$ long videos ($2\thinspace500$ frames on average). Moreover, we report results on the UAV123~\cite{UAV123} dataset, consisting of $123$ videos which feature small targets and distractor objects. Results are also reported on the 30 FPS version of the need for speed (NFS)~\cite{NFS} dataset, containing $100$ videos with fast motions. Finally, we evaluate on the 100 videos of OTB-100~\cite{OTB100}. Our tracker is evaluated in terms of overlap precision (OP). For a threshold $T \in [0, 1]$, OP$_T$ is the percentage of frames in which the IoU overlap between the estimated and ground truth target bounding box is larger than $T$. By averaging OP$_T$ over $T \in [0, 1]$, the \textit{AUC} score is then obtained. For TrackingNet, the term \textit{Success} is used in place of AUC. Results in terms of AUC on all five datasets are found in Table~\ref{tab:tracking}. To ensure significance, the average AUC over $5$ runs is reported for our trackers. We observe that DiMP-NCE+ consistently outperforms both our DiMP-KLD-IS baseline, and a variant employing NCE instead of NCE+. Compared to previous approaches, only the very recent SiamRCNN~\cite{voigtlaender2020siam} achieves results competitive with our DiMP-NCE+. SiamRCNN is however slower than DiMP-NCE+ (5~FPS vs 30 FPS) and employs a larger backbone network (ResNet101 vs ResNet50). Results for the ResNet50 version of SiamRCNN are only available on two of the datasets, on which it is outperformed by our DiMP-NCE+. More detailed results are provided in the supplementary material.

\begin{table}[t]
\centering
\resizebox{1.0\textwidth}{!}{%
\begin{tabular}{l@{~~~}c@{~~~}c@{~~~}c@{~~~}c@{~~~}c@{~~~}c@{~~~}c@{~~~}c@{~~~}c@{~~~}c@{~~~}c}
	\toprule
	&MDNet &UPDT &DaSiamRPN &ATOM &SiamRPN++ &DiMP &SiamRCNN &PrDiMP &DiMP- &DiMP- &\textbf{DiMP-}\\
	&\cite{MDNet} &\cite{UPDT} &\cite{DaSiamRPN} &\cite{danelljan2019atom} &\cite{SiamRPN++} &\cite{bhat2019learning} &\cite{voigtlaender2020siam} &\cite{danelljan2020probabilistic} &KLD-IS &NCE &\textbf{NCE+}\\
	\midrule
	
	\multirow{1}{18mm}{TrackingNet}
	&60.6 &61.1 &63.8 &70.3 &73.3 &74.0 &\textbf{81.2} &75.8 &78.1 &77.1 &78.7\\
	
	\multirow{1}{15mm}{LaSOT}
	&39.7 &- &- &51.5 &49.6 &56.9 &\textbf{64.8} (62.3) &59.8 &63.1 &62.8 &63.7\\
	
	\multirow{1}{15mm}{UAV123}
	&52.8 &54.5 &57.7 &63.2 &61.3 &64.3 &64.9 &66.7 &66.6 &65.2 &\textbf{67.2}\\

	\multirow{1}{15mm}{NFS}
	&42.2 &53.7 &- &58.4 &- &62.0 &63.9 &63.5 &64.7 &64.3 &\textbf{65.0}\\
	
	\multirow{1}{15mm}{OTB-100}
	&67.8 &70.2 &65.8 &66.9 &69.6 &68.4 &70.1 (68.0) &69.6 &70.1 &69.3 &\textbf{70.7}\\
	
	\bottomrule
\end{tabular}

	
	
	

	
	

	}\vspace{-3.5mm}
\caption{Results for the visual tracking experiments. The AUC (Success) metric is reported on five common datasets. Our proposed DiMP-NCE+ tracker significantly outperforms strong baselines and achieves state-of-the-art performance on all five datasets. For SiamRCNN~\cite{voigtlaender2020siam}, results for the ResNet50 version are given in parentheses when available.}\vspace{-3.0mm}
	\label{tab:tracking}
\end{table}

\section{Conclusion}
\label{section: conclusion}
We proposed a simple yet highly effective extension of NCE to train EBMs $p(y | x; \theta)$ for computer vision regression tasks. Our proposed method NCE+ can be understood as a direct generalization of NCE, accounting for noise in the annotation process of real-world datasets. We also provided a detailed comparison of NCE+ and six popular methods from literature, the results of which suggest that NCE+ should be considered the go-to training method. This comparison is the first comprehensive study of how EBMs should be trained for best possible regression performance. Finally, we applied our proposed NCE+ to the task of visual tracking, achieving state-of-the-art performance on five commonly used datasets. We hope that our simple training method and promising results will encourage the research community to further explore the application of EBMs to various regression tasks.

\parsection{Acknowledgments}
This research was financially supported by the Swedish Foundation for Strategic Research via the project \emph{ASSEMBLE}, the Swedish Research Council via the project \emph{Learning flexible models for nonlinear dynamics}, the ETH Z\"urich Fund (OK), a Huawei Technologies Oy (Finland) project, an Amazon AWS grant, and Nvidia.

\bibliography{references}

\clearpage

\renewcommand{\thefigure}{S\arabic{figure}}
\setcounter{figure}{0}

\renewcommand{\thetable}{S\arabic{table}}
\setcounter{table}{0}

\renewcommand{\thealgorithm}{S\arabic{algorithm}}
\setcounter{algorithm}{0}

\renewcommand{\theequation}{S\arabic{equation}}
\setcounter{equation}{0}


\section*{\centering{\\Supplementary Material}}

In this supplementary material, we provide additional details and results. It consists of Appendix~\ref{appendix:prediciton} - Appendix~\ref{appendix:visual_tracking}. Appendix~\ref{appendix:prediciton} contains a detailed algorithm for our employed prediction strategy. Further experimental details are provided in Appendix~\ref{appendix:1dregression} for 1D regression, and in Appendix~\ref{appendix:object_detection} for object detection. Lastly, Appendix~\ref{appendix:visual_tracking} contains details and further results for the visual tracking experiments. Note that equations, tables, figures and algorithms in this supplementary document are numbered with the prefix "S". Numbers without this prefix refer to the main paper.

\appendix
\begin{appendices}
\section{Prediction Algorithm}
\label{appendix:prediciton}
Our prediction procedure (Section 2.2) is detailed in Algorithm~\ref{algo:prediction}, where $\lambda$ denotes the gradient ascent step-length, $\eta$ is a decay of the step-length and $T$ is the number of iterations.

\begin{algorithm}
\caption{Prediction via gradient-based refinement.}
\label{algo:prediction}
\textbf{Input:} $x^\star$, $\hat{y}$, $T$, $\lambda$, $\eta$.
\begin{algorithmic}[1]
    \State $y \gets \hat{y}$.
    \For{\texttt{$t = 1, \dots, T$}}
        \State \texttt{PrevValue} $\gets$ $f_{\theta}(x^\star, y)$.
        \State $\Tilde{y} \gets y + \lambda \nabla_{y} f_{\theta}(x^\star, y)$.
        \State \texttt{NewValue} $\gets$ $f_{\theta}(x^\star, \Tilde{y})$.
        \If { $\texttt{NewValue} > \texttt{PrevValue}$}
            \State $y \gets \Tilde{y}$.
        \Else
            \State $\lambda \gets \eta \lambda$.
        \EndIf
    \EndFor
    \State \textbf{Return} $y$.
\end{algorithmic}
\end{algorithm}\vspace{-3mm}
\section{1D Regression}
\label{appendix:1dregression}
Here, we provide details on the two synthetic datasets, the network architecture, the evaluation procedure, and hyperparameters used for our 1D regression experiments (Section 4.1). For all seven training methods, the DNN $f_{\theta}(x, y)$ was trained (by minimizing the associated loss $J(\theta)$) for $75$ epochs with a batch size of $32$ using the ADAM~\cite{kingma2014adam} optimizer.

\subsection{Datasets}
\begin{figure}[t]%
    \begin{minipage}{0.3125\textwidth}%
        \centering
        \includegraphics[width=1.0\textwidth]{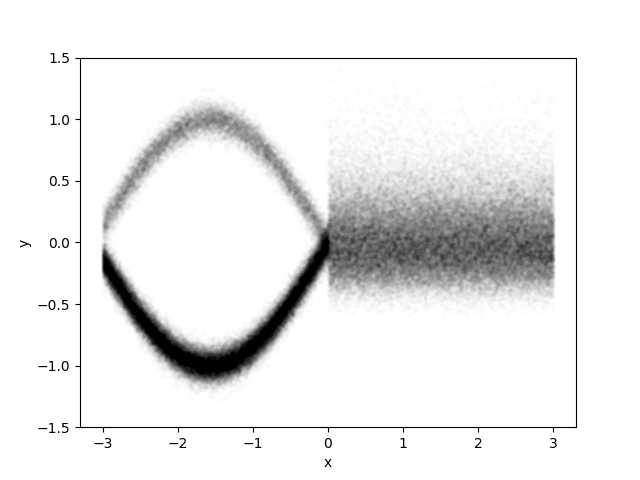}\vspace{-7.0mm}
        \caption{Visualization of the true $p(y | x)$ for the first 1D regression dataset.}\vspace{-3mm}
        \label{fig:1dregression_1_gt}%
    \end{minipage}
    \quad
    \begin{minipage}{0.3125\textwidth}%
        \centering
        \includegraphics[width=1.0\textwidth]{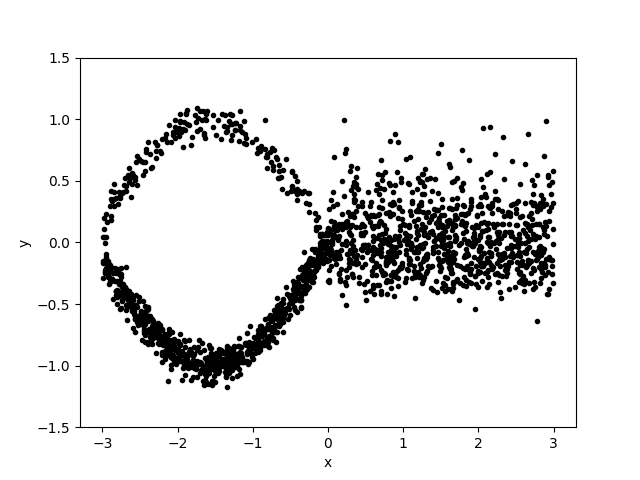}\vspace{-7.0mm}
        \caption{Training data $\{(x_i, y_i)\}_{i=1}^{2000}$ for the first 1D regression dataset.}\vspace{-3mm}
        \label{fig:1dregression_1_data}%
    \end{minipage}%
    \quad
    \begin{minipage}{0.3125\textwidth}%
        \centering
        \includegraphics[width=1.0\textwidth]{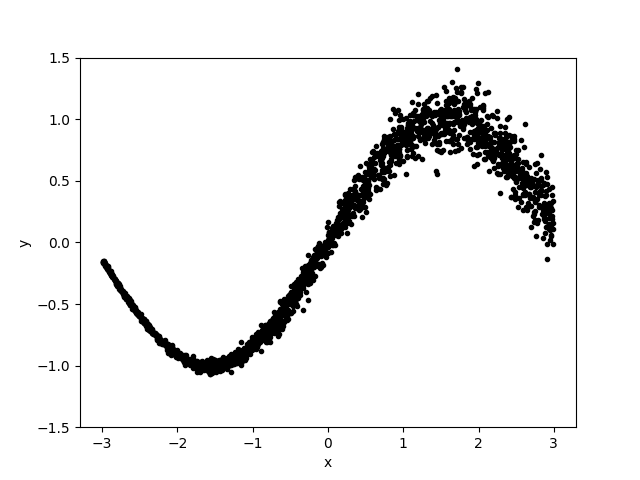}\vspace{-7.0mm}
        \caption{Training data $\{(x_i, y_i)\}_{i=1}^{2000}$ for the second 1D regression dataset.}\vspace{-3mm}
        \label{fig:1dregression_2_data}%
    \end{minipage}%
\end{figure}

The ground truth $p(y | x)$ for the first dataset is visualized in Figure~\ref{fig:1dregression_1_gt}. It is defined by a mixture of two Gaussian components (with weights $0.2$ and $0.8$) for $x < 0$, and a log-normal distribution (with $\mu = 0.0$, $\sigma= 0.25$) for $x \geq 0$. The training data $\mathcal{D}_1 = \{(x_i, y_i)\}_{i=1}^{2000}$ was generated by uniform random sampling of $x$ in the interval $[-3, 3]$, and is visualized in Figure~\ref{fig:1dregression_1_data}. The ground truth $p(y | x)$ for the second dataset is defined according to,
\begin{equation}
    \begin{gathered}
        p(y | x) = \mathcal{N}\big(y; \mu(x), \sigma^2(x)\big),\\
        \mu(x) = \sin(x), \quad \sigma(x) = 0.15(1 + e^{-x})^{-1}.
    \end{gathered}
    \label{eq:1dregression_2}
\end{equation}
The training data $\mathcal{D}_2 = \{(x_i, y_i)\}_{i=1}^{2000}$ was generated by uniform random sampling of $x$ in the interval $[-3, 3]$, and is visualized in Figure~\ref{fig:1dregression_2_data}.

\subsection{Network Architecture}
The DNN $f_\theta(x, y)$ is a feed-forward network taking $x \in \mathbb{R}$ and $y \in \mathbb{R}$ as inputs. It consists of two fully-connected layers (dimensions: $1 \rightarrow 10$, $10 \rightarrow 10$) for $x$, one fully-connected layer ($1 \rightarrow 10$) for $y$, and four fully-connected layers ($20 \rightarrow 10$, $10 \rightarrow 10$, $10 \rightarrow 10$, $10 \rightarrow 1$) processing the concatenated $(x, y)$ feature vector.

\subsection{Evaluation}
\label{section:1dregression_evaluation}
The training methods are evaluated in terms of the KL divergence $D_\mathrm{KL}(p(y | x) \parallel p(y | x; \theta))$ between the learned EBM $p(y | x; \theta) = e^{f_{\theta}(x, y)}/\int e^{f_{\theta}(x, \tilde{y})} d\tilde{y}$ and the true conditional density $p(y | x)$. To approximate $D_\mathrm{KL}(p(y | x) \parallel p(y | x; \theta))$, we compute $e^{f_{\theta}(x, y)}$ and $p(y | x)$ for all $(x, y)$ pairs in a $2048 \times 2048$ uniform grid in the region $\{(x, y) \in \mathbb{R}^{2}: x \in [-3, 3], y \in [-3, 3]\}$. We then normalize across all values associated with each $x$, employ the formula for KL divergence between two discrete distributions $q_1(y)$ and $q_2(y)$,
\begin{equation}
    D_\mathrm{KL}(q_1 \parallel q_2) = \sum_{y \in \mathcal{Y}}^{} q_1(y) \log \frac{q_1(y)}{q_2(y)},
\end{equation}
and finally average over all $2048$ values of $x$. For each dataset and training method, we independently train the DNN $f_\theta(x, y)$ and compute $D_\mathrm{KL}(p(y | x) \parallel p(y | x; \theta))$ $20$ times. We then take the mean of the $5$ best runs, and finally average this value for the two datasets.

\subsection{Hyperparameters}
The number of samples $M=1024$ for all applicable training methods. All other hyperparameters were selected to optimize the performance, evaluated according to Section~\ref{section:1dregression_evaluation}.

\parsection{ML-IS}
Following \cite{gustafsson2019learning}, we set $K=2$ in the proposal distribution $q(y | y_i)$ in (4). After ablation, we set $\sigma_1 = 0.2$, $\sigma_2 = 1.6$.

\parsection{KLD-IS}
We use the same proposal distribution $q(y | y_i)$ as for ML-IS. After ablation, we set $\sigma = 0.025$ in $p(y | y_i) = \mathcal{N}(y; y_i, \sigma^2 I)$.

\parsection{ML-MCMC}
After ablation, we set the Langevin dynamics step-length $\alpha = 0.05$.

\parsection{NCE}
To match ML-IS, we set $K=2$ in the noise distribution $p_N(y | y_i)$ in (11). After ablation, we set $\sigma_1 = 0.1$, $\sigma_2 = 0.8$.

\parsection{DSM}
After ablation, we set $\sigma = 0.2$ in $p_\sigma(\tilde{y} | y_i) = \mathcal{N}(\tilde{y}; y_i, \sigma^2 I)$.

\parsection{NCE+}
We use the same noise distribution $p_N(y | y_i)$ as for NCE. After ablation, we set $\beta = 0.025$.

\subsection{Qualitative Results}
\begin{figure}[t]%
    \begin{minipage}{0.3125\textwidth}%
        \centering
        \includegraphics[width=1.0\textwidth]{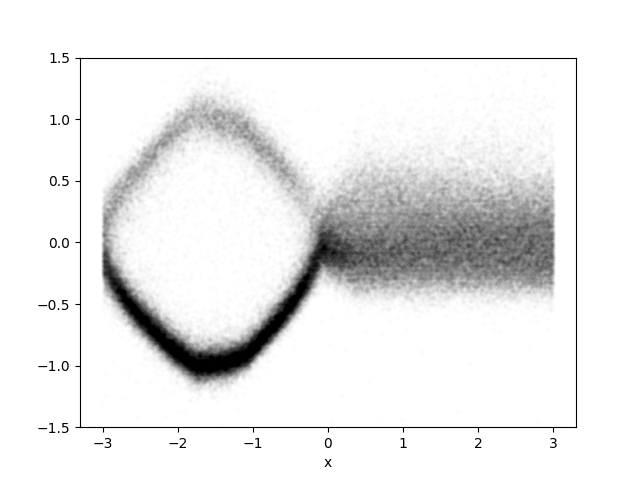}\vspace{-8.0mm}
        \caption{Example of $p(y | x; \theta)$ trained with NCE.}\vspace{-0mm}
        \label{fig:1dregression_1_example_nce}%
    \end{minipage}
    \quad
    \begin{minipage}{0.3125\textwidth}%
        \centering
        \includegraphics[width=1.0\textwidth]{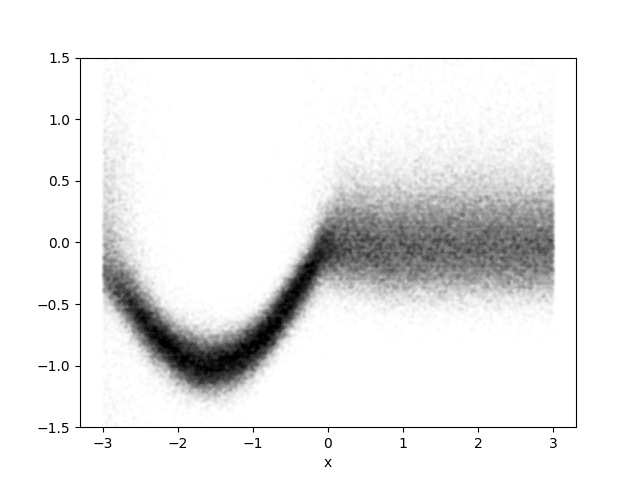}\vspace{-8.0mm}
        \caption{Example of $p(y | x; \theta)$ trained with DSM.}\vspace{-0mm}
        \label{fig:1dregression_1_example_dsm}%
    \end{minipage}%
    \quad
    \begin{minipage}{0.3125\textwidth}%
        \centering
        \includegraphics[width=1.0\textwidth]{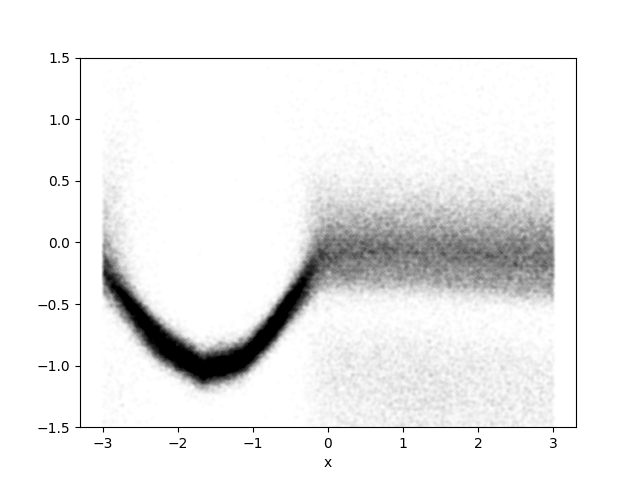}\vspace{-8.0mm}
        \caption{Example of $p(y | x; \theta)$ trained with SM.}\vspace{-0mm}
        \label{fig:1dregression_1_example_sm}%
    \end{minipage}%
\end{figure}

An example of $p(y | x; \theta)$ trained using NCE on the first dataset is visualized in Figure~\ref{fig:1dregression_1_example_nce}. As can be observed, this is quite close to the true $p(y | x)$ visualized in Figure~\ref{fig:1dregression_1_gt}. Similar results are obtained with all four top-performing training methods. Examples of $p(y | x; \theta)$ instead trained using DSM and SM are visualized in Figure~\ref{fig:1dregression_1_example_dsm} and Figure~\ref{fig:1dregression_1_example_sm}, respectively. These do not approximate the true $p(y | x)$ quite as well, matching the worse performance in terms of $D_\mathrm{KL}$ reported in Table 1.

\begin{table}[t]
\centering
\resizebox{1.0\textwidth}{!}{%
		\begin{tabular}{l@{\hspace{0.5cm}}cccccccc}
\toprule
 &ML-IS &ML-MCMC-1 &ML-MCMC-4 &ML-MCMC-8 &KLD-IS &NCE& DSM &\textbf{NCE+}\\
\midrule
$\lambda_{\mathrm{pos}}$             &0.0004 &0.000025 &0.000025 &0.000025 &0.0004 &0.0004 &0.000025 &0.0008\\

$\lambda_{\mathrm{size}}$            &0.0016 &0.0001 &0.0001 &0.0001 &0.0016 &0.0016 &0.0001 &0.0032\\
\bottomrule
\end{tabular}

	}\vspace{-3.5mm}
\caption{Used step-lengths $\lambda_{\mathrm{pos}}$ and $\lambda_{\mathrm{size}}$ for the object detection experiments.}\vspace{-4.0mm}
	\label{tab:comparison_object_detection_step}
\end{table}

\section{Object Detection}
\label{appendix:object_detection}
Here, we provide details on the prediction procedure and hyperparameters used for our object detection experiments (Section 4.2). We employ an identical network architecture and training procedure as described in \cite{gustafsson2019learning}, only modifying the loss when using a different method than ML-IS to train $f_\theta(x, y)$.

\subsection{Prediction}
Predictions $y^{\star}$ are produced by performing guided NMS~\cite{jiang2018acquisition} followed by gradient-based refinement (Algorithm~\ref{algo:prediction}), taking the Faster-RCNN detections as initial estimates $\hat{y}$. As in \cite{gustafsson2019learning}, we run $T=10$ gradient ascent iterations. We fix the step-length decay to $\eta = 0.5$, which is the value used in \cite{gustafsson2019learning}. For each trained model, we select the gradient ascent step-length $\lambda$ to optimize performance in terms of AP on the \textit{2017 val} split of COCO~\cite{lin2014microsoft}. Like \cite{gustafsson2019learning}, we use different step-lengths for the bounding box position ($\lambda_{\mathrm{pos}}$) and size ($\lambda_{\mathrm{size}}$). We start this ablation with $\lambda_{\mathrm{pos}} = 0.0001$, $\lambda_{\mathrm{size}} = 0.0004$. The used step-lengths for all training methods are given in Table~\ref{tab:comparison_object_detection_step}.

\subsection{Hyperparameters}
The number of samples $M=128$ for all applicable training methods. All other hyperparameters were selected to optimize performance in terms of AP on the \textit{2017 val} split of COCO~\cite{lin2014microsoft}.

\parsection{ML-IS}
Following \cite{gustafsson2019learning}, we set $K=3$ in the proposal distribution $q(y | y_i)$ in (4) with $\sigma_1 = 0.0375$, $\sigma_2 = 0.075$, $\sigma_3 = 0.15$.

\parsection{KLD-IS}
We use the same proposal distribution $q(y | y_i)$ as for ML-IS. Based on the ablation study in Table~\ref{tab:detection_ablation_kld-is}, we set $\sigma = 0.0225$ in $p(y | y_i) = \mathcal{N}(y; y_i, \sigma^2 I)$.

\begin{table}
    \begin{minipage}{0.475\textwidth}%
        \centering
        \resizebox{1.0\textwidth}{!}{%
        		\begin{tabular}{l@{\hspace{0.5cm}}ccccc}
\toprule
$\sigma$                       &0.0075 &0.015 &0.0225 &0.03 &0.0375\\
\midrule
AP (\%) $\uparrow$             &38.32 &39.19 &\textbf{39.38} &39.33 &39.23\\
\bottomrule
\end{tabular}
        	}
        \caption{Ablation study for KLD-IS, on the \textit{2017 val} split of COCO~\cite{lin2014microsoft}.}\vspace{-1.0mm}
        	\label{tab:detection_ablation_kld-is}
    \end{minipage}
    \quad
    \begin{minipage}{0.475\textwidth}%
        \centering
        \resizebox{0.825\textwidth}{!}{%
        		\begin{tabular}{l@{\hspace{0.5cm}}ccc}
\toprule
$\alpha$                       &0.000001 &0.00001 &0.0001\\
\midrule
AP (\%) $\uparrow$             &36.14 &\textbf{36.19} &36.04\\
\bottomrule
\end{tabular}
        	}
        \caption{Ablation study for ML-MCMC-1, on the \textit{2017 val} split of COCO~\cite{lin2014microsoft}.}\vspace{-1.0mm}
        	\label{tab:detection_ablation_ml-mcmc}
    \end{minipage}%
\end{table}

\parsection{ML-MCMC}
Based on the ablation study in Table~\ref{tab:detection_ablation_ml-mcmc}, we set the Langevin dynamics step-length $\alpha = 0.00001$.

\parsection{NCE}
To match ML-IS, we set $K=3$ in the noise distribution $p_N(y | y_i)$ in (11). Based on the ablation study in Table~\ref{tab:detection_ablation_nce}, we set $\sigma_1 = 0.075$, $\sigma_2 = 0.15$, $\sigma_3 = 0.3$.

\begin{table}
\centering
\resizebox{1.0\textwidth}{!}{%
		\begin{tabular}{l@{\hspace{0.5cm}}ccccc}
\toprule
$\{\sigma_{k}\}_{k=1}^{3}$                       &\{0.0125, 0.025, 0.05\} &\{0.025, 0.05, 0.1\} &\{0.05, 0.1, 0.2\} &\{0.075, 0.15, 0.3\} &\{0.1, 0.2, 0.4\}\\
\midrule
AP (\%) $\uparrow$            &38.58 &38.95 &39.12 &\textbf{39.17} &39.05\\
\bottomrule
\end{tabular}
	}\vspace{-3.5mm}
\caption{Ablation study for NCE, on the \textit{2017 val} split of COCO~\cite{lin2014microsoft}.}\vspace{-3.0mm}
	\label{tab:detection_ablation_nce}
\end{table}

\parsection{DSM}
Based on the ablation study in Table~\ref{tab:detection_ablation_dsm}, we set $\sigma = 0.075$ in $p_\sigma(\tilde{y} | y_i) = \mathcal{N}(\tilde{y}; y_i, \sigma^2 I)$.

\begin{table}
    \begin{minipage}{0.475\textwidth}%
        \centering
        \resizebox{0.805\textwidth}{!}{%
        		\begin{tabular}{l@{\hspace{0.5cm}}ccc}
\toprule
$\sigma$                       &0.0375 &0.075 &0.15\\
\midrule
AP (\%) $\uparrow$             &36.11 &\textbf{36.12} &36.05\\
\bottomrule
\end{tabular}
        	}
        \caption{Ablation study for DSM, on the \textit{2017 val} split of COCO~\cite{lin2014microsoft}.}\vspace{-0.0mm}
        	\label{tab:detection_ablation_dsm}
    \end{minipage}
    \quad
    \begin{minipage}{0.475\textwidth}%
        \centering
        \resizebox{0.805\textwidth}{!}{%
        		\begin{tabular}{l@{\hspace{0.5cm}}ccc}
\toprule
$\beta$                       &0.05 &0.1 &0.15\\
\midrule
AP (\%) $\uparrow$            &39.27 &\textbf{39.36} &39.32\\
\bottomrule
\end{tabular}
        	}
        \caption{Ablation study for NCE+, on the \textit{2017 val} split of COCO~\cite{lin2014microsoft}.}\vspace{-0.0mm}
        	\label{tab:detection_ablation_nce+}
    \end{minipage}%
\end{table}

\parsection{NCE+}
We use the same noise distribution $p_N(y | y_i)$ as for NCE. Based on the ablation study in Table~\ref{tab:detection_ablation_nce+}, we set $\beta = 0.1$.

\subsection{Detailed Results}
A comparison of the training methods on the \textit{2017 val} split of COCO~\cite{lin2014microsoft} is provided in Table~\ref{tab:comparison_object_detection_val}.

\begin{table}[t]
\centering
\resizebox{1.0\textwidth}{!}{%
		\begin{tabular}{l@{\hspace{0.5cm}}cccccccc}
\toprule
 &ML-IS &ML-MCMC-1 &ML-MCMC-4 &ML-MCMC-8 &KLD-IS &NCE& DSM &\textbf{NCE+}\\
\midrule
AP (\%) $\uparrow$             &39.11 &36.19 &36.24 &36.25 &\textbf{39.38} &39.17 &36.12 &39.36\\
AP$_\text{50} (\%)$ $\uparrow$ &57.95 &57.34 &57.45 &57.28 &\textbf{58.07} &57.96 &57.29 &57.99\\
AP$_\text{75} (\%)$ $\uparrow$ &41.97 &38.77 &38.81 &38.88 &42.47 &42.07 &38.84 &\textbf{42.63}\\
Training Cost $\downarrow$     &1.03 &2.47 &7.05 &13.3 &\textbf{1.02} &1.04 &3.84 &1.09\\
\bottomrule
\end{tabular}
	}\vspace{-3.5mm}
\caption{Comparison of training methods for the object detection experiments, on the \textit{2017 val} split of COCO~\cite{lin2014microsoft}. NCE+ and KLD-IS achieve the best performance.}\vspace{-1.0mm}
	\label{tab:comparison_object_detection_val}
\end{table}
\begin{table}[b]
\centering
\resizebox{1.0\textwidth}{!}{%
	
	
	
	

\begin{tabular}{l@{~~~}c@{~~~}c@{~~~}c@{~~~}c@{~~~}c@{~~~}c@{~~~}c@{~~~}c@{~~~}c@{~~~}c@{~~~}c@{~~~}c}
	\toprule
	&SiamFC &MDNet &UPDT &DaSiamRPN &ATOM &SiamRPN++ &DiMP &SiamRCNN &PrDiMP &DiMP- &DiMP- &\textbf{DiMP-}\\
	&\cite{SiameseFC} &\cite{MDNet} &\cite{UPDT} &\cite{DaSiamRPN} &\cite{danelljan2019atom} &\cite{SiamRPN++} &\cite{bhat2019learning} &\cite{voigtlaender2020siam} &\cite{danelljan2020probabilistic} &KLD-IS &NCE &\textbf{NCE+}\\
	\midrule
	
	\multirow{1}{25mm}{Precision $\uparrow$}
	&53.3 &56.5 &55.7 &59.1 &64.8 &69.4 &68.7 &\textbf{80.0} &70.4 &73.3 &69.8 &73.7\\
	
	\multirow{1}{25mm}{Norm. Prec. $\uparrow$}
	&66.6 &70.5 &70.2 &73.3 &77.1 &80.0 &80.1 &\textbf{85.4} &81.6 &83.5 &82.4 &83.7\\
	
	\multirow{1}{25mm}{Success (AUC) $\uparrow$}
	&57.1 &60.6 &61.1 &63.8 &70.3 &73.3 &74.0 &\textbf{81.2} &75.8 &78.1 &77.1 &78.7\\
	
	\bottomrule
\end{tabular}
	}\vspace{-3.5mm}
\caption{Full results on the TrackingNet~\cite{muller2018trackingnet} test set, in terms of precision, normalized precision, and success (AUC). Our proposed DiMP-NCE+ is here only outperformed by the very recent SiamRCNN~\cite{voigtlaender2020siam}. SiamRCNN is however slower than DiMP-NCE+ (5~FPS vs 30 FPS) and employs a larger backbone network (ResNet101 vs ResNet50).}\vspace{-0.0mm}
	\label{tab:tracking_trackingnet}
\end{table}

\section{Visual Tracking}
\label{appendix:visual_tracking}
Here, we provide detailed results and hyperparameters for our visual tracking experiments (Section 5). We employ an identical network architecture, training procedure and prediction procedure for DiMP-KLD-IS, DiMP-NCE and DiMP-NCE+, only the loss is modified.

\subsection{Training Parameters}
DiMP-KLD-IS is obtained by combining the DiMP~\cite{bhat2019learning} method for center point regression with the PrDiMP~\cite{danelljan2020probabilistic} bounding box regression approach, and modifying a few training parameters. Specifically, we change the batch size from $10$ to $20$, we change the LaSOT sampling weight from $0.25$ to $1.0$, we change the number of samples per epoch from $26\thinspace000$ to $40\thinspace000$, and we add random horizontal flipping with probability $0.5$. Since we increase the batch size, we also freeze conv1, layer1 and layer2 of the ResNet backbone to save memory.

\begin{figure}[t]%
    \begin{minipage}{0.475\textwidth}%
        \centering
        \includegraphics[width=1.0\textwidth]{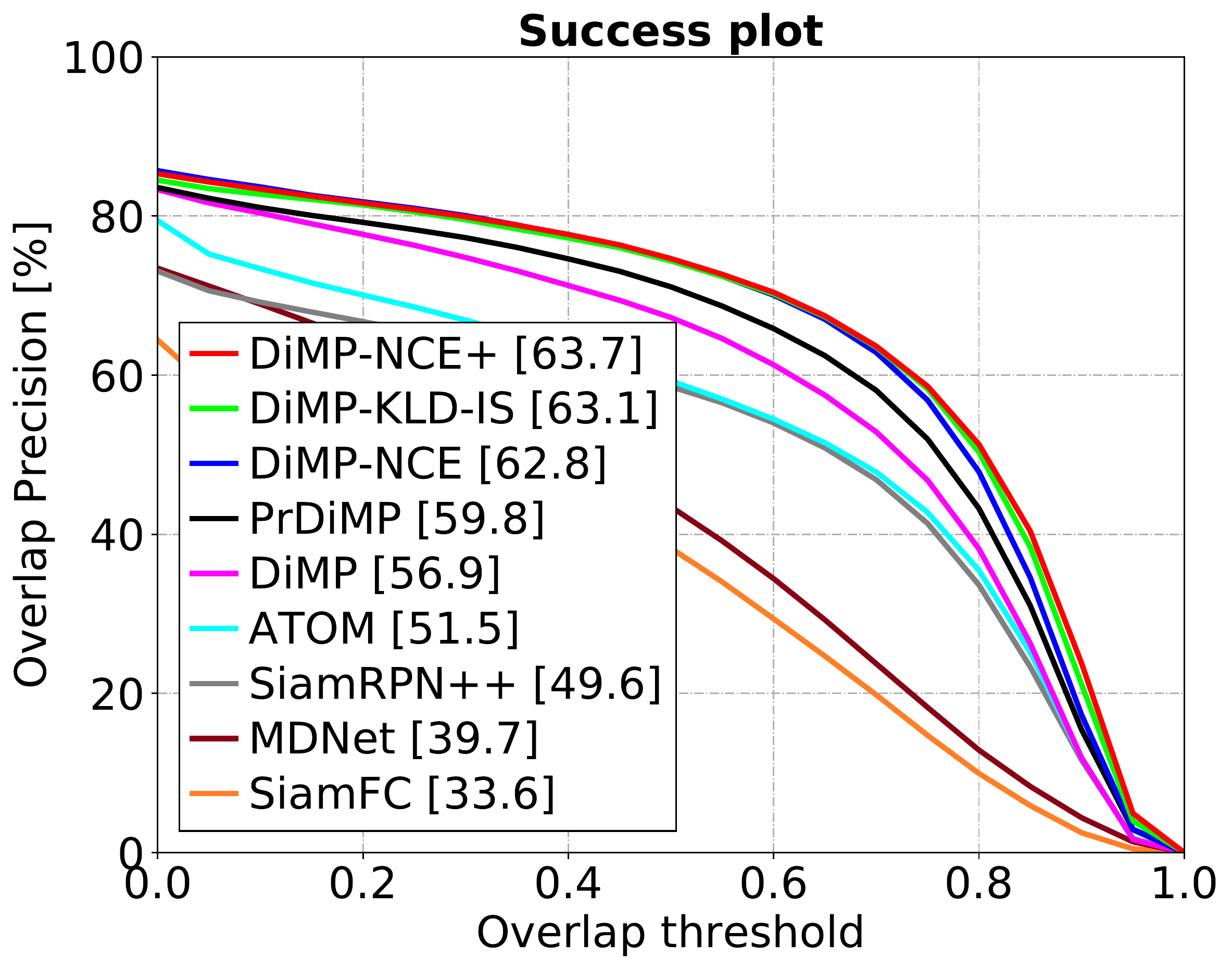}\vspace{-7.0mm}
        \caption{Success plot on LaSOT~\cite{fan2019lasot}.}\vspace{-3mm}
        \label{fig:tracking_lasot}%
    \end{minipage}
    \quad
    \begin{minipage}{0.475\textwidth}%
        \centering
        \includegraphics[width=1.0\textwidth]{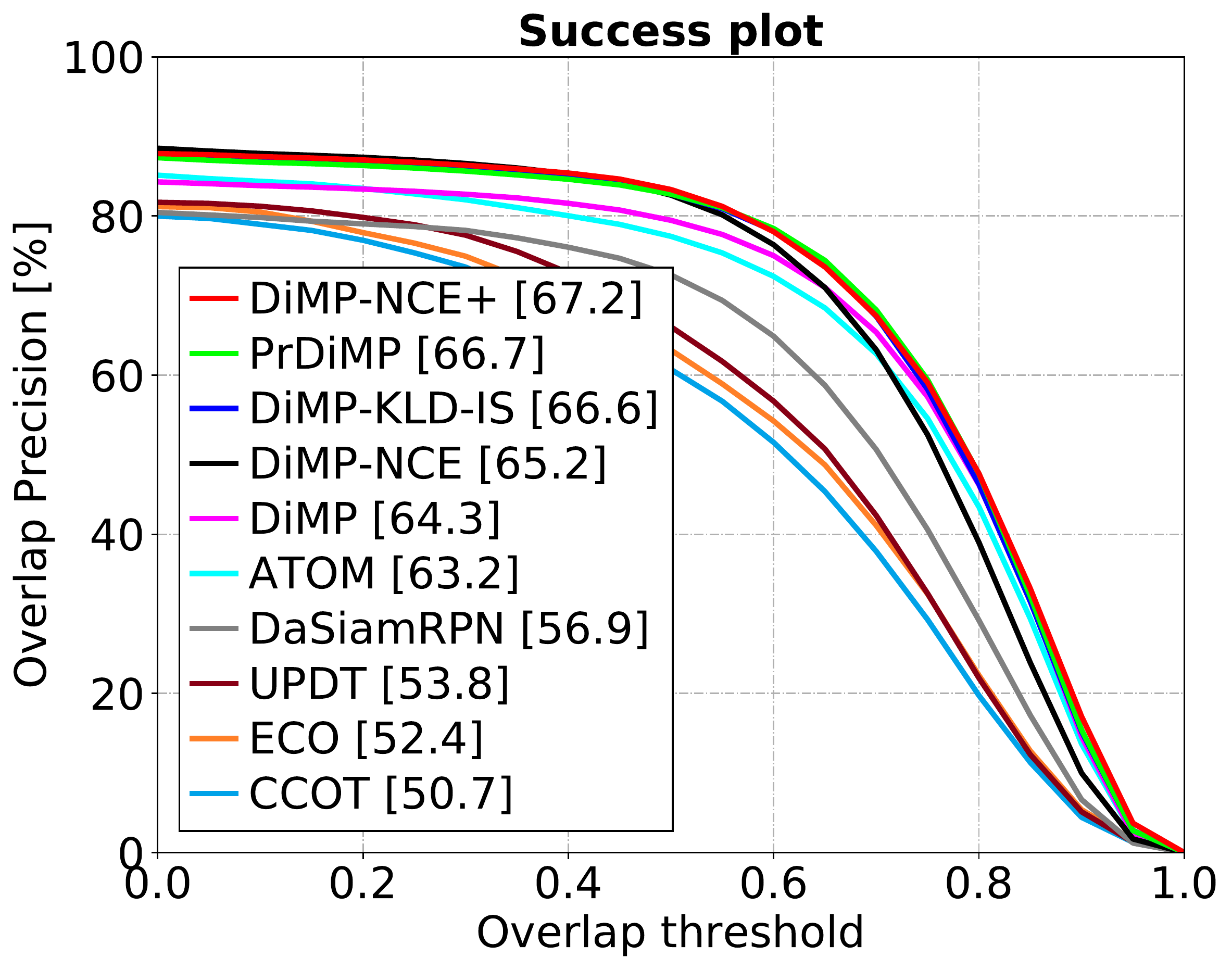}\vspace{-7.0mm}
        \caption{Success plot on UAV123~\cite{UAV123}.}\vspace{-3mm}
        \label{fig:tracking_uav123}%
    \end{minipage}%
\end{figure}%

\begin{figure}[t]%
    \begin{minipage}{0.475\textwidth}%
        \centering
        \includegraphics[width=1.0\textwidth]{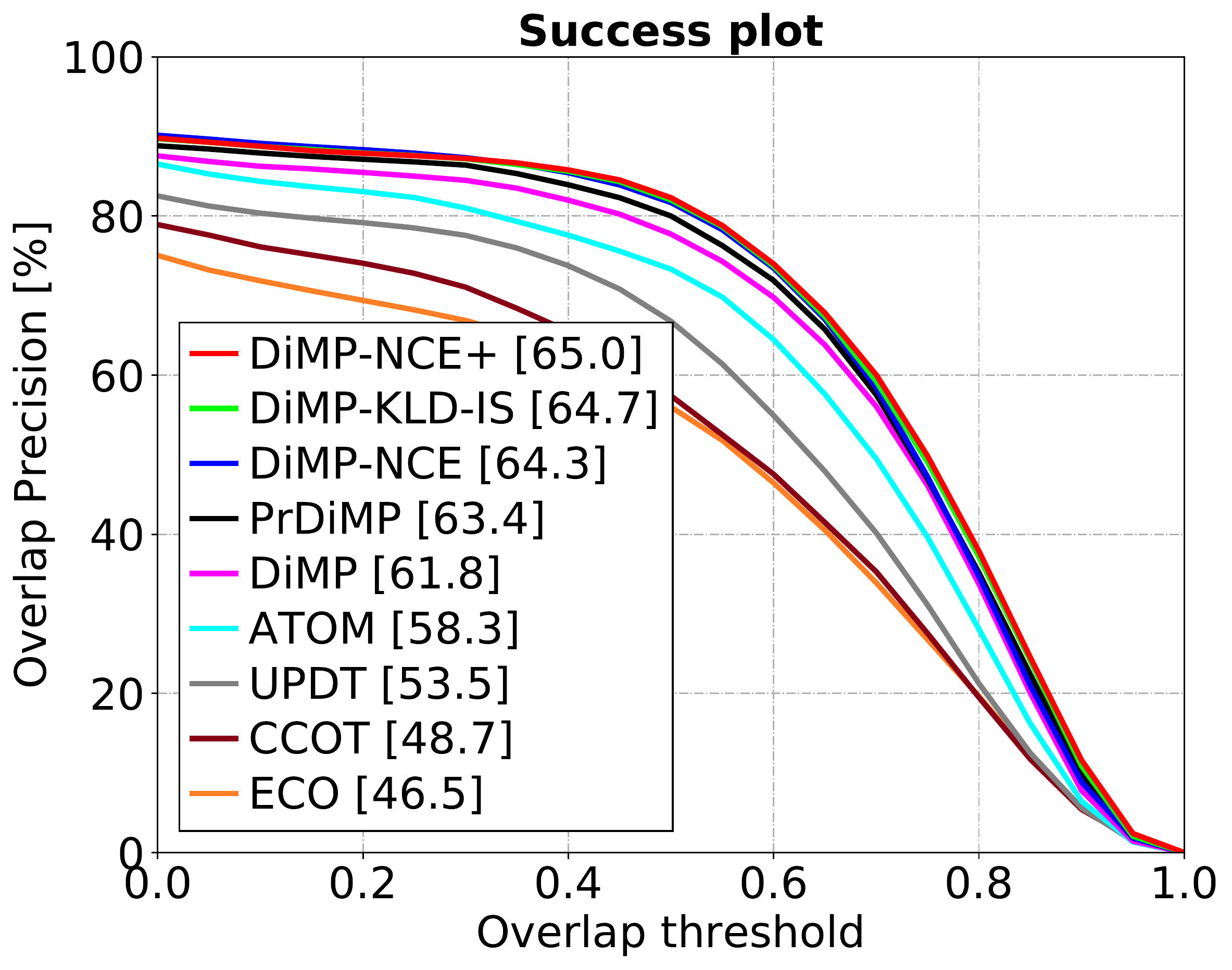}\vspace{-7.0mm}
        \caption{Success plot on NFS~\cite{NFS}.}\vspace{-3mm}
        \label{fig:tracking_nfs}%
    \end{minipage}
    \quad
    \begin{minipage}{0.475\textwidth}%
        \centering
        \includegraphics[width=1.0\textwidth]{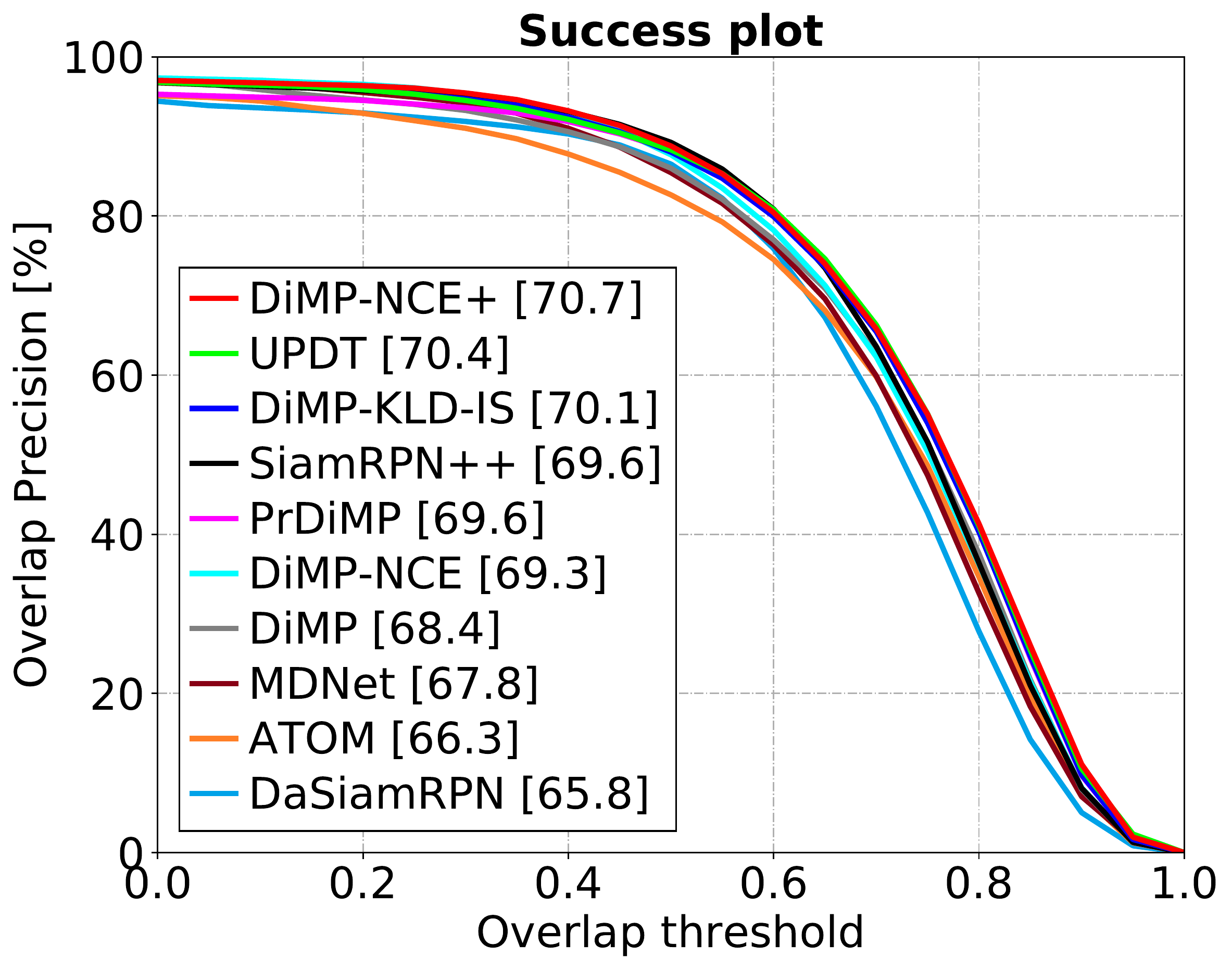}\vspace{-7.0mm}
        \caption{Success plot on OTB-100~\cite{OTB100}.}\vspace{-3mm}
        \label{fig:tracking_otb100}%
    \end{minipage}%
\end{figure}%

\subsection{Hyperparameters}
The number of samples $M=128$ for all three training methods.

\parsection{DiMP-KLD-IS}
Following PrDiMP, we set $K=2$ in the proposal distribution $q(y | y_i)$ in (4) with $\sigma_1 = 0.05$, $\sigma_2 = 0.5$, and we set $\sigma = 0.05$ in $p(y | y_i) = \mathcal{N}(y; y_i, \sigma^2 I)$.

\parsection{DiMP-NCE}
Matching DiMP-KLD-IS, we set $K=2$ in the noise distribution $p_N(y | y_i)$ in (11) with $\sigma_1 = 0.05$, $\sigma_2 = 0.5$. A quick ablation study on the validation set of GOT-10k~\cite{huang2019got} did not find values of $\sigma_1, \sigma_2$ resulting in improved performance.

\parsection{DiMP-NCE+}
We use the same noise distribution $p_N(y | y_i)$ as for NCE. We set $\beta=0.1$, as this corresponded to the best performance on the object detection experiments (Table~\ref{tab:detection_ablation_nce+}).

\subsection{Detailed Results}
Full results on the TrackingNet~\cite{muller2018trackingnet} test set, in terms of all three TrackingNet metrics, are found in Table~\ref{tab:tracking_trackingnet}. Success plots for LaSOT, UAV123, NFS and OTB-100 are found in Figure~\ref{fig:tracking_lasot}-\ref{fig:tracking_otb100}, showing the overlap precision OP$_T$ as a function of the overlap threshold $T$.

\end{appendices}

\end{document}